\documentclass[sigconf]{acmart}
\usepackage{algorithm}
\usepackage{algorithmic}
\AtBeginDocument{%
  }

\usepackage{ulem}
\usepackage{graphicx}
\usepackage{amsmath}
\usepackage{booktabs} 
\usepackage{multirow} 
\usepackage{booktabs} 
\usepackage{xcolor}
\usepackage{tabularx}
\usepackage{graphicx} 
\usepackage{multirow} 
\usepackage{subfigure}
\usepackage{xcolor}
\usepackage{colortbl}
\usepackage{enumitem}

\renewcommand\footnotetextcopyrightpermission[1]{} 
\begin{document}

\title{Mitigating Group-Level Fairness Disparities in Federated Visual Language Models}


\author{Chaomeng Chen}
\affiliation{%
  \institution{Great Bay University}
  \institution{Tsinghua Shenzhen International Graduate School, Tsinghua University}
  \country{}
}

\author{Zitong Yu * }\thanks{ * Corresponding author}
\affiliation{%
  \institution{Great Bay University}
  \country{}
}

\author{Junhao Dong}
\affiliation{%
  \institution{Nanyang Technological University}
  \country{}
}

\author{Sen Su}
\affiliation{%
  \institution{Beijing University of Posts and Telecommunications}
  \country{}
}

\author{Linlin Shen}
\affiliation{%
  \institution{Shenzhen University}
  \country{}
}

\author{Shutao Xia}
\affiliation{%
  \institution{Tsinghua Shenzhen International Graduate School, Tsinghua University}
  \institution{Pengcheng Laboratory}
  \country{}
}

\author{Xiaochun Cao}
\affiliation{%
  \institution{Shenzhen Campus of Sun Yat-sen University}
  \country{}
}


\begin{abstract}
Visual language models (VLMs) have shown remarkable capabilities in multimodal tasks but face challenges in maintaining fairness across demographic groups, particularly when deployed in federated learning (FL) environments. This paper addresses the critical issue of group fairness in federated VLMs by introducing FVL-FP, a novel framework that combines FL with fair prompt tuning techniques. We focus on mitigating demographic biases while preserving model performance through three innovative components: (1) Cross-Layer Demographic Fair Prompting (CDFP), which adjusts potentially biased embeddings through counterfactual regularization; (2) Demographic Subspace Orthogonal Projection (DSOP), which removes demographic bias in image representations by mapping fair prompt text to group subspaces; and (3) Fair-aware Prompt Fusion (FPF), which dynamically balances client contributions based on both performance and fairness metrics. Extensive evaluations across four benchmark datasets demonstrate that our approach reduces demographic disparity by an average of 45\% compared to standard FL approaches, while maintaining task performance within 6\% of state-of-the-art results. FVL-FP effectively addresses the challenges of non-IID data distributions in federated settings and introduces minimal computational overhead while providing significant fairness benefits. Our work presents a parameter-efficient solution to the critical challenge of ensuring equitable performance across demographic groups in privacy-preserving multimodal systems.
\end{abstract}




\maketitle

\section{Introduction}
Visual language models (VLMs), such as CLIP \cite{radford2021learning} and BLIP \cite{li2022blip}, have recently demonstrated significant capabilities in multimodal AI applications, including image understanding, visual reasoning, and cross-modal generation \cite{alayrac2022flamingo, jia2021scaling, yu2022coca}. These models leverage contrastive learning between image and text modalities, enabling powerful zero-shot capabilities across diverse tasks. Through pre-training on large-scale datasets with billions of image-text pairs, VLMs have established new benchmarks in multimodal understanding. However, these models now face the challenge of increasingly stringent global data privacy regulations \cite{gdpr2016,ccpa2020}, which restrict the centralized collection and processing of user data. Federated Learning (FL) \cite{yang2019federated, mcmahan2017communication} offers an effective alternative to enable collaboration among distributed nodes while protecting data privacy, introducing new synergies when combined with VLMs. By keeping sensitive data localized while sharing only model updates, FL contributes to developing privacy-preserving multimodal systems.

\begin{figure}[t]
\vspace{-0.8em}

	\centering
	\includegraphics[width=0.9\linewidth]{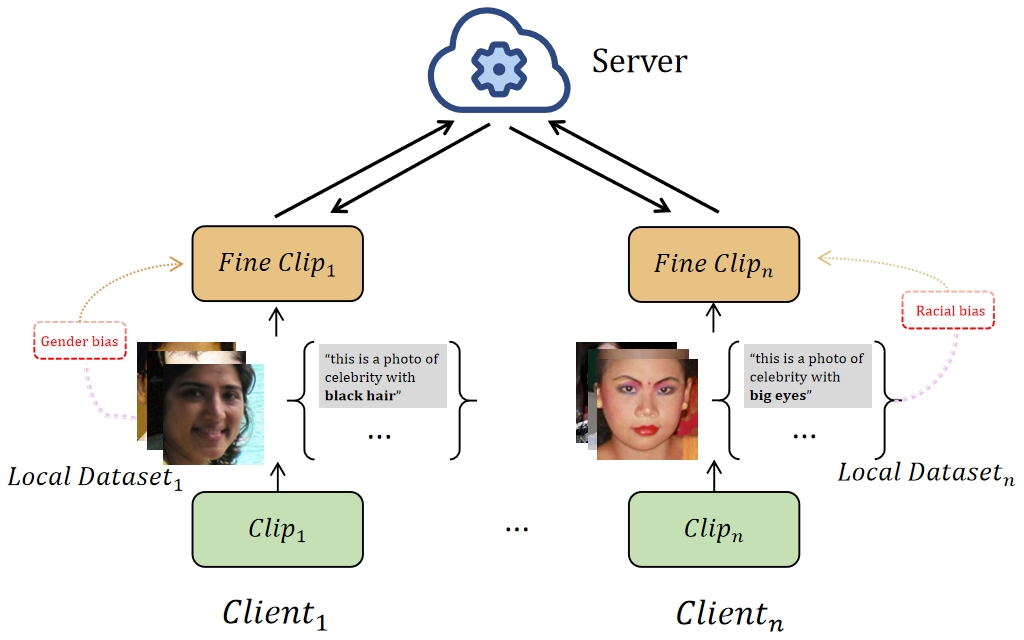}
\vspace{-1.6em}
	\caption{Federated Vision-Language Models. For each node, clients fine-tune local vision-language models based on local datasets. However, local datasets contain underlying biases in group fairness, which affect global group fairness. In this work, we propose a federated vision-language model framework that eliminates bias using fairness prompt tuning.} 
	\label{intro}
    \vspace{-2.0em}
\end{figure}

Despite existing methods in this emerging field, the federated visual language models (FL-VLMs) paradigm often involves fine-tuning and training on local datasets that reflect regional or demographic characteristics. This process may introduce or amplify biases inherent to specific demographic groups, such as different genders, races, age groups, and socioeconomic backgrounds, subsequently affecting group fairness \cite{de2021stereotype,wang2022investigating,zhao2021captioning}. For instance, recent studies have shown that VLMs may associate certain professions predominantly with specific genders or ethnicities, or generate descriptions that reinforce harmful stereotypes \cite{wang2022revise, zhao2018gender}, as shown in Figure \ref{intro}. When deployed in federated environments, these biases can become more pronounced due to the heterogeneous nature of client data distributions, creating systematic disparities in model performance across different demographic groups.

Addressing bias is a fundamental prerequisite\textemdash rather than an afterthought\textemdash for developing responsible AI systems. Existing strategies to mitigate bias in FL or VLMs, such as data augmentation \cite{zhao2018gender,zhang2022towards}, adversarial debiasing \cite{zhang2020towards}, and incorporating fairness constraints during local model retraining \cite{2109-08604,chuang2023debiasing}, are still in their early stages and face significant challenges. These challenges include: (1) high computational costs associated with training, making retraining VLMs with billions of network parameters on distributed nodes with limited computational resources impractical; (2) significant differences in data distribution across devices, making it complex to achieve global group fairness without affecting the local model performance; and (3) the tension between optimizing for task performance and fairness metrics, which often involves significant trade-offs that are exacerbated in federated settings where client objectives may differ a lot.

Recent advances in prompt tuning \cite{jia2022visual, zhou2022learning, zhou2022conditional} have demonstrated its effectiveness as a minimalist yet powerful technique for training VLMs with significantly reduced parameter updates. By optimizing only a small set of continuous prompt vectors rather than the entire parameter space, prompt tuning achieves comparable performance to full fine-tuning while requiring orders of magnitude fewer trainable parameters. Building on this foundation, subsequent research \cite{lu2023federated,zhang2023learning} has successfully integrated prompt tuning into FL environments with VLMs, facilitating model updates through prompt exchange while significantly reducing communication overhead. This approach is particularly promising for resource-constrained devices, as it minimizes both computational requirements and network bandwidth usage.

In response to these technical advances and persistent challenges, we propose a pioneering framework, dubbed Federated Visual Language Models with Fair Prompt Tuning (FVL-FP), specifically designed to mitigate group fairness issues in FL-VLMs. Our research addresses the critical gap between federated VLM optimization and fairness considerations, offering a parameter-efficient approach that maintains standard performance while enhancing equality across multiple demographic groups. The FVL-FP framework is built around three innovative components:

\begin{itemize}
\item The Cross-Layer Demographic Fair Prompting (CDFP) algorithm adjusts potentially biased embeddings to generate fair prompt embeddings. This enhancement aims to improve the fairness of model parameters by identifying and neutralizing bias directions in the embedding space through counterfactual regularization. CDFP operates locally on each client, adapting to the specific bias patterns presented in local data distributions while maintaining a unified fairness objective.

\item The Demographic Subspace Orthogonal Projection (DSOP) algorithm removes gender and demographic bias in image representations by mapping fair text prompts to group subspaces. By constructing orthogonal projections that separate protected attribute information from semantic content, DSOP ensures that model predictions do not rely on sensitive characteristics. This geometric approach thus provides an interpretable mechanism for debiasing that preserves the rich representational capabilities of VLMs.

\item The Fair-aware Prompt Fusion (FPF) algorithm dynamically adjusts the weights of these fair prompts across clients to ensure the stability of global prompt updates throughout the training process, striking a balance between performance and fairness. FPF incorporates client-specific fairness metrics into the aggregation process, prioritizing contributions from clients that demonstrate both strong task performance and equitable outcomes across demographic groups.
\end{itemize}

Through rigorous evaluation of four benchmark datasets, FVL-FP has been proven to achieve substantial group fairness across various VLM tasks, despite challenges posed by different levels of data heterogeneity. Our extensive experiments demonstrate that FVL-FP reduces demographic disparity by an average of 45\% compared to standard FL approaches while maintaining task performance within $\pm6\%$ of state-of-the-art results.

\begin{figure*}[htbp]
\vspace{-0.9em}
\centering
\includegraphics[width=0.9\linewidth]{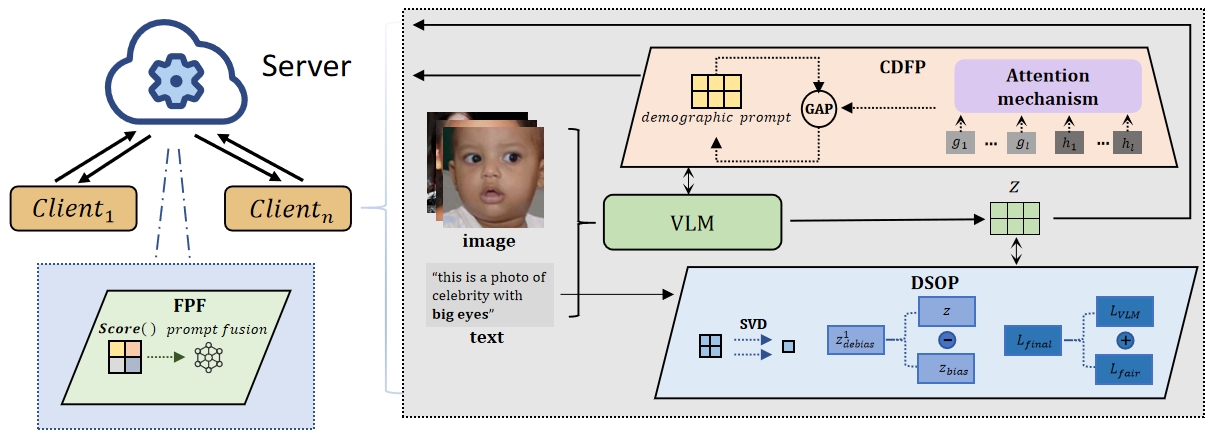}
\vspace{-1.5em}
\caption{The framework of FVL-FP. For each node, clients optimize for fairness through CDFP and DSOP algorithms. Following the local training phase, clients transmit local fairness prompts to the central server, which then performs fairness prompt aggregation using the FPF algorithm to construct a global fair prompt.}
\label{FVL-FP}
\vspace{-0.9em}
\end{figure*}

\section{Related Work}

\subsection{Group Fairness in Visual Language Models}
As VLMs have been widely applied across various domains, concerns about group fairness biases have increasingly grown. Research in this field has primarily focused on identifying and quantifying gender, racial, and other biases in VLMs, achieving significant progress. Innovative quantitative metrics \cite{de2021stereotype,wong2019time,webster2018mind} have provided deeper insights into model biases, establishing a theoretically solid foundation for debiasing efforts. Techniques to enhance model fairness through increased dropout regularization \cite{webster2020measuring} have proven effective in reducing gender bias in visual representations without impairing model performance, mainly by mitigating the model's dependency on gender-specific features. Additionally, Counterfactual Data Augmentation (CDA) \cite{zhao2018gender} has emerged as an effective strategy, utilizing gender attribute swapping and other attribute word modifications in image-text pairs to balance datasets and reduce biases in visual-language representations \cite{barikeri2021redditbias}. The GEEP approach \cite{fatemi2021improving} has pioneered in enhancing fairness by creating neutral visual datasets and subsequently fine-tuning the model, offering a novel data preprocessing method to mitigate multimodal biases. The ADEPT algorithm \cite{yang2023adept} enhances the fairness of large VLMs through stream learning and debiasing criteria. The Iterative Nullspace Projection (INLP) method \cite{ravfogel2020null} has also been used to eliminate linear correlations between visual-text embeddings and protected attributes, providing a robust theoretical framework for addressing biases in VLMs. Furthermore, the Self-Debias technique \cite{schick2021self} represents a significant post-hoc multimodal generation debiasing method, utilizing probabilistic adjustments between biased and unbiased visual-text content to achieve debiasing, demonstrating the potential for bias reduction through model post-processing.
Despite these successes, many methods still require extensive retraining, leading to significant resource consumption, extended training periods, and risks of catastrophic forgetting, which pose challenges for practical applications. Our research explores more efficient and practical debiasing strategies to address these challenges faced in group fairness of VLMs.

\vspace{-0.6em}
\subsection{Group Fairness in Federated Learning}
FL is a prominent distributed machine learning framework, particularly valued for its ability to train models collaboratively across decentralized nodes while preserving the privacy of underlying data. A pivotal concern within this framework is the assurance of equitable outcomes across varied demographic groups—including gender, ethnicity, and age—which has become a focal point in recent scholarly discussions \cite{DBLP:conf/sdm/DuXWT21, DBLP:journals/corr/abs-2109-08344}.
Innovations in this domain have introduced new fairness constraints and optimization techniques aimed at enhancing group fairness. Notable advancements include the application of advanced differential multiplier methods \cite{2109-08604} and the implementation of debiasing mechanisms like FairBatch, which integrates server-side weight adjustments \cite{roh2020fairbatch}. Additionally, the principle of max-min demographic fairness has been rigorously applied to improve fairness metrics in FL settings, promoting equal treatment across the most and least advantaged groups \cite{papadaki2022minimax}.
To tackle the challenges of heterogeneous group fairness in FL, recent works have proposed local debiasing techniques alongside global weighted aggregation strategies \cite{ezzeldin2023fairfed}. The adoption of Secure Multi-party Computation (SMC) techniques further enhances privacy protections in these scenarios \cite{DBLP:journals/corr/abs-2205-11584}. However, these methods often require frequent retraining at the device level, leading to substantial computational and communication overhead, especially when integrated with VLMs.
Our research introduces a novel, lightweight debiasing algorithm specifically designed for group fairness in FL. This algorithm aims to provide fair outcomes without the extensive computational and communicational demands typical of previous methods, thereby significantly boosting the practicality and applicability of FL-VLMs.

\vspace{-1.0em}
\section{Problem Formulation}

Existing bias mitigation approaches predominantly rely on computationally intensive techniques such as data augmentation \cite{zhao2018gender,zhang2022towards}, adversarial debiasing \cite{zhang2020towards}, and fairness-constrained retraining \cite{2109-08604,chuang2023debiasing}. These methodologies face significant limitations in FL-VLM contexts due to (1) the prohibitive computational costs associated with retraining billion-parameter VLMs on resource-constrained devices, (2) the inherent heterogeneity in data distributions across federated participants, and (3) the fundamental tension between optimizing for task performance and fairness metrics, which is further exacerbated in federated settings where client objectives may vary considerably.

Recent advances in prompt tuning \cite{jia2022visual, zhou2022learning, zhou2022conditional} present an opportunity to address these challenges through parameter-efficient adaptation of VLMs. By optimizing only a small set of continuous prompt vectors rather than the entire model, prompt tuning achieves comparable performance to full fine-tuning while requiring orders of magnitude fewer trainable parameters. This approach has been successfully extended to federated environments \cite{lu2023federated,zhang2023learning}, facilitating model updates through prompt exchange while significantly reducing communication overhead.

Thus, we formally define the problem of group fairness in FL-VLMs within this prompt tuning paradigm as follows:

\textbf{Federated Visual Language Model Setting:} Consider a federation of $N$ clients, where each client $i \in \{1, 2, \ldots, N\}$ possesses a local dataset $D_i = \{(x_j^i, y_j^i, g_j^i)\}_{j=1}^{|D_i|}$. Here, $x_j^i$ represents a multimodal input (image-text pair), $y_j^i$ denotes the corresponding label, and $g_j^i \in \{g, h\}$ indicates the sensitive attribute (e.g., gender) for the $j$-th sample in client $i$'s dataset. The objective is to collaboratively train a global VLM $f_{\theta}(\cdot)$ with network parameters $\theta$ while ensuring both high standard performance associated with group fairness.

\textbf{Group Fairness in Federated Visual Language Models:} We defined the global group fairness of the federated visual language model, using the equal opportunity difference (EOD) metric as an example, which measures the difference in true positivity rates between sensitive attribute groups:
\begin{equation}\small
	\begin{aligned}
		F_{global} = \left| \frac{1}{N} \sum_{i=1}^N \Pr(\hat{Y}_i = 1 | G_i = g, Y_i = 1) \right. \\
		\left. - \frac{1}{N} \sum_{i=1}^N \Pr(\hat{Y}_i = 1 | G_i = h, Y_i = 1) \right|,
	\end{aligned}
\end{equation}
where $\hat{Y}_i$ represents predicted outcomes on client $i$'s data. $G_i$ denotes the sensitive attribute, and $Y_i$ indicates the ground truth. This formulation quantifies the absolute difference in average true positive rates between demographic groups across all clients in the federation. A lower value of $F_{global}$ indicates more equitable treatment across groups, with perfect fairness achieved at $F_{global} = 0$.

\textbf{Fair Prompt Tuning in Federated Settings:} Instead of fine-tuning the entire VLM, we focus on optimizing a small set of continuous prompt vectors $P = \{p_1, p_2, ..., p_m\}$ that are prepended to the input text embeddings. Given a pre-trained VLM $f_{\theta}(\cdot)$ with frozen parameters $\theta$, each client $i$ locally optimizes its prompt parameters $P_i$ on dataset $D_i$ to minimize both task-specific loss and fairness disparity. Specifically, the objective function for client $i$ balances performance and fairness:
\begin{equation}\small
\min_{P_i} \mathcal{L}_{task}(f_{\theta}(P_i, D_i)) + \lambda \mathcal{L}_{fair}(f_{\theta}(P_i, D_i)),
\end{equation}
where $\mathcal{L}_{task}$ represents the task-specific loss (e.g., cross-entropy for classification), $\mathcal{L}_{fair}$ quantifies the fairness violation, and $\lambda$ controls the trade-off between task performance and fairness.

\section{Methology}
In this section, we introduce the FVL-FP framework, which dynamically adjusts training prompts to maintain fairness within each node through the LFPT algorithm, and ensures group fairness among multiple nodes via the FPA algorithm at the server. The preliminaries are summarized in Appendix A.

\subsection{Overview of FVL-FP Framework}

We propose FVL-FP, a novel framework that enhances the fairness of FL-VLMs. As illustrated in Figure~\ref{FVL-FP}, our framework consists of three key algorithmic components: 1) Cross-Layer Demographic Fair Prompting (CDFP), which trains demographic-aware soft prompts on the client side to capture group-specific characteristics and mitigate biases present in each demographic group; 2) Demographic Subspace Orthogonal Projection (DSOP), which identifies and projects away unfair directions in the representation space to reduce unwanted correlations with protected demographic attributes while maintaining semantic meaningfulness; and 3) Fair-aware Prompt Fusion (FPF), which operates on the server side to aggregate locally trained prompts with a novel weighting mechanism that prioritizes fairness alongside accuracy. These components work in concert through an iterative process where clients first use CDFP to tune prompts locally, then apply DSOP to ensure fairness constraints, after which the server employs FPF to fuse these prompts into a globally fair representation, thereby leveraging diverse demographic information while maintaining privacy and achieving significant improvements in fairness metrics.

\vspace{-0.9em}
\subsection{Cross-Layer Demographic Fair Prompting}

To mitigate bias in local VLMs, we implement debiasing through prompt tuning. Specifically, we propose a Cross-Layer Demographic Fair Prompting (CDFP) algorithm that effectively suppresses demographically correlated signals in VLM features while preserving the original model's performance.

\begin{figure}[t]
\vspace{-0.9em}
\centering
\includegraphics[width=\linewidth]{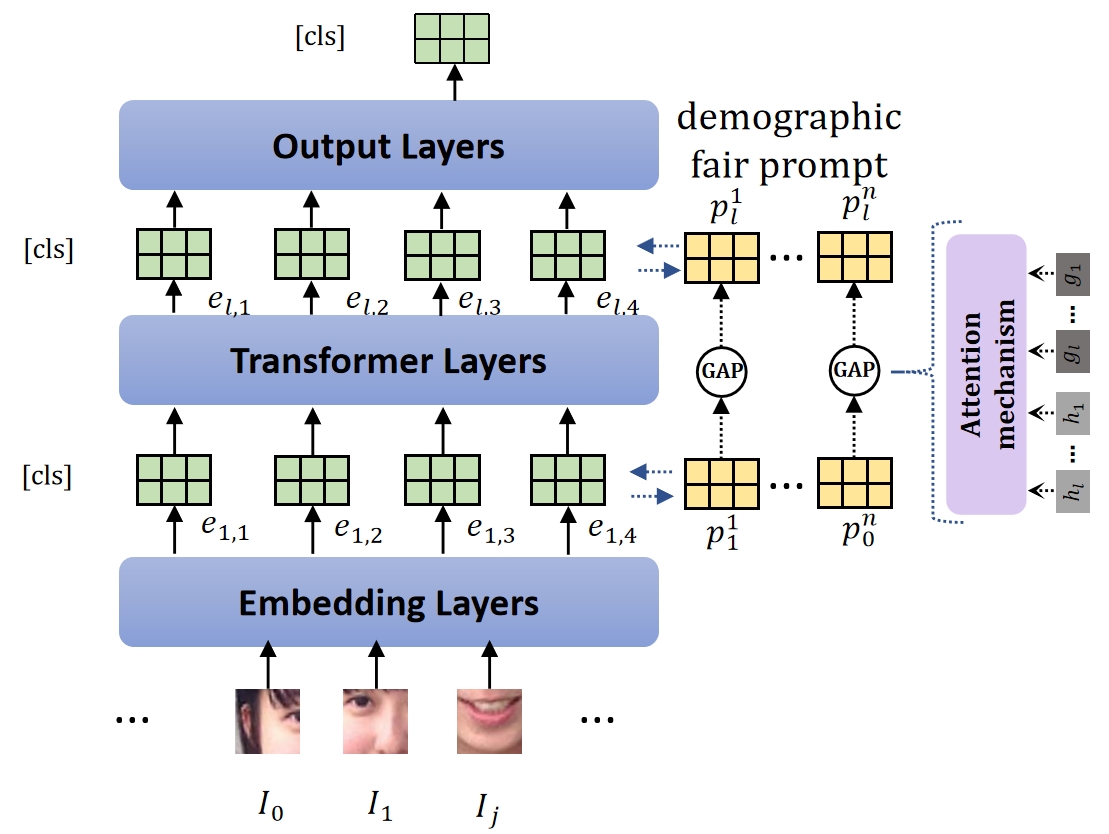}
\vspace{-2.3em}
\caption{Cross-Layer Demographic Fair Prompting Algorithm. Our method inserts demographic fair prompts at the embedding layer and propagates them through transformer layers with adaptive dynamic residual connections. The GAP mechanism enables learnable cross-layer connections for effective bias mitigation while preserving model performance.}
\vspace{-0.9em}
\label{CDFP}
\end{figure}

In the CDFP algorithm(Figure \ref{CDFP}), we first decompose the VLM's image encoder $f_1$ into $L$ sequential layers. At the input layer, the image $\mathbf{x}$ is partitioned into $J$ fixed-size patches ${I_1, I_2, \dots, I_J}$, each of size $h \times w$. These patches are embedded at layer 0 as follows:
\begin{equation}\small
e_{0,j} = \text{Embed}(I_j), \quad e_{0,j} \in \mathbb{R}^d, \quad j \in {1, 2, \dots, J}.
\end{equation}

These initial embeddings then propagate through multiple Transformer layers of the image encoder. At the $l$-th layer ($l \in {1,2,\dots,L}$), the transformation can be expressed as:
\begin{equation}\small
[e_{l,0}, E_l] = f_1^{\mathrm{transformer}} \bigl( [e_{l-1,0}; E_{l-1}] \bigr),
\end{equation}
where $E_l = [e_{l,1}, e_{l,2}, \dots, e_{l,J}]$ represents the features of all image patches at layer $l$. The final output vector $\mathbf{e}_{L,0}$ (corresponding to the [CLS] token) serves as the global representation of the image.

To mitigate VLM's inherent bias toward demographic attributes, our approach strategically inserts visual prompt vectors at both the embedding layer and subsequent Transformer layers. Unlike conventional prompt tuning methods, we introduce a \textbf{demographic fair prompt} $\mathcal{P}_0 = {p_0^1, p_0^2, ..., p_0^K} \in \mathbb{R}^{K \times d}$, where each of the $K$ basis vectors represent different sensitive group categories (such as gender, race, age). This sensitive group fair prompt is inserted at layer 0 as follows:
\begin{equation}\small
[\mathbf{e}_{0,0}, \ \underbrace{p_0^1, \dots, p_0^K}_{\mathcal{P}_0},\ \mathbf{e}_{0,1}, \dots, \mathbf{e}_{0,J}],
\end{equation}

For subsequent layers, we propose an adaptive dynamic residual connection mechanism. Rather than simply transforming the prompt vectors independently at each layer, we establish learnable connections between prompt vectors across different layers to enhance fairness control. Specifically, the transformation at the $l$-th layer can be represented as:
\begin{equation}\small
[e_{l,0},\mathcal{P}_l, E_l] = f_1^{\mathrm{transformer}} \bigl( [e_{l-1,0};\mathcal{P}_{l-1}; E_{l-1}] \bigr),
\end{equation}
where the fairness prompt at layer $l$ is further refined through our adaptive dynamic residual connection with lower-layer prompts:
\begin{equation}\small
\mathcal{P}_l = \mathcal{P}_l + \text{GAP}_l(\mathcal{P}_{<l}).
\end{equation}

Here, $\text{GAP}_l$ is a layer-specific Gated Attention Pooling function:
\begin{equation}\small
\text{GAP}_l(\mathcal{P}_{<l}) = \sum_{i=0}^{l-1} \gamma_{l,i} \cdot \mathcal{P}_i,
\end{equation}
where $\gamma_{l,i}$ are attention weights computed through a learnable attention mechanism instead of fixed hyperparameters:
\begin{equation}\small
\gamma_{l,i} = \frac{\exp(g_l^T \cdot h_i)}{\sum_{j=0}^{l-1} \exp(g_l^T \cdot h_j)}.
\end{equation}

In this formulation, $g_l$ is a learnable query vector for layer $l$, and $h_i$ is the contextualized representation of the prompt at layer $i$. This improvement allows the model to automatically learn the optimal connection strengths between different layers without manual hyperparameter tuning.

Through this Cross-Layer Demographic Fair Prompting and adaptive cross-layer prompt sharing mechanism, we can effectively suppress the model's excessive attention to demographic attributes while maintaining its performance on downstream tasks. This method not only simplifies the implementation of fairness control but also provides a more flexible and interpretable approach to balance the model's fairness and utility.

\vspace{-0.6em}
\subsection{Demographic Subspace Orthogonal Projection}

\begin{figure}[t]
\vspace{-1.0em}
    \centering
    \includegraphics[width=\linewidth]{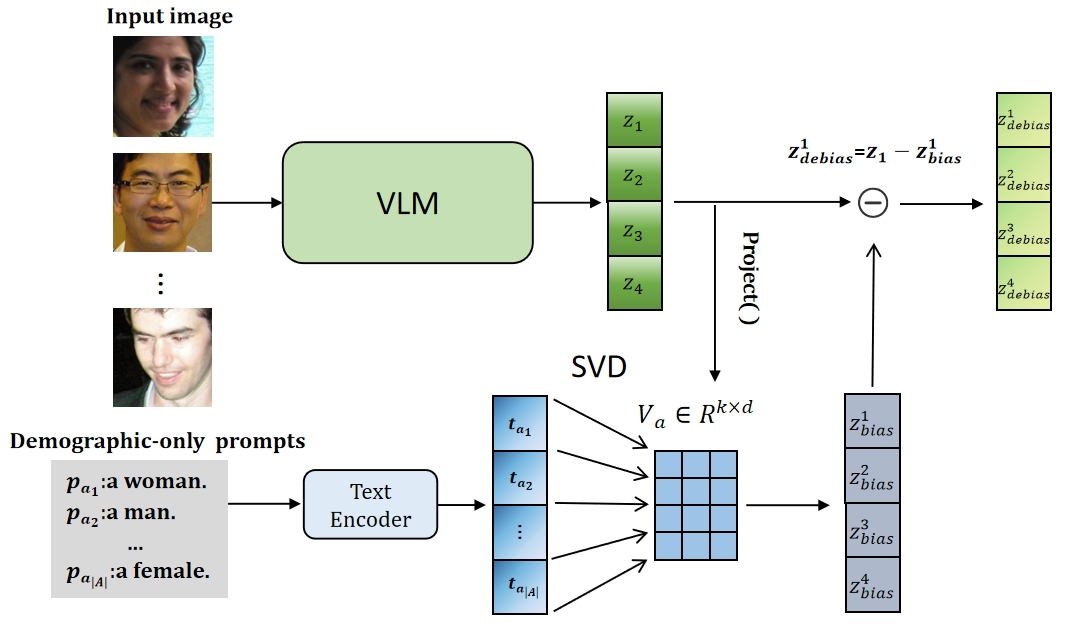}
    \vspace{-1.8em}
    \caption{Demographic Subspace Orthogonal Projection. We orthogonally project visual representations away from demographic subspaces to reduce bias while preserving task-relevant information.}
    \label{DSOP}
    \vspace{-1.0em}
\end{figure}

To enhance the fairness of VLM representations, we propose a demographic subspace orthogonal projection approach that systematically removes demographic-related components from the visual representation $z$. By constructing a demographic subspace and projecting out the corresponding components orthogonally, we enable VLM to become invariant to sensitive demographic attributes (e.g., gender, race, and age) while preserving task-relevant semantic information. Figure \ref{DSOP} describes the operation of the DSOP in detail.

\subsubsection{Demographic Subspace Construction}
We begin by constructing a set of demographic-specific prompts $\{p_{a_1}, \ldots, p_{a_{|A|}}\}$ that explicitly describe different values of a demographic attribute $a$ (e.g., ``a photo of a man'', ``a photo of a woman''). These prompts are encoded through the VLM text encoder $f_r$ to obtain a set of text embeddings $T_a = \{t_{a_1}, \ldots, t_{a_{|A|}}\}$, where $t_{a_i} = f_r(p_{a_i})$. We then organize these vectors into a matrix $T_a \in \mathbb{R}^{|A| \times d}$, where $d$ represents the embedding dimension. By applying Singular Value Decomposition (SVD), we extract the top-$k$ principal directions that collectively span the demographic subspace $V_a \in \mathbb{R}^{k \times d}$.

\subsubsection{Debiasing via Orthogonal Projection}
For an input image $x$ with its prompted visual representation $z$, we project $z$ onto the demographic subspace $V_a$ to identify its demographic component $z_{\text{bias}} = \text{Proj}_{V_a}(z)$. The debiased representation is then obtained by subtracting this demographic component:
\begin{equation}\small
    z_{\text{debiased}} = z - z_{\text{bias}}
\end{equation}

This orthogonal projection ensures that $z_{\text{debiased}}$ is minimally influenced by the demographic attributes represented in subspace $V_a$.

\subsubsection{Fairness-aware Contrastive Learning}
To further suppress residual demographic signals, we introduce a fairness-aware contrastive loss. This loss penalizes high cosine similarity between the normalized debiased representation $\tilde{z}_{\text{debiased}}$ and any demographic prompt embedding. Formally, we define:
\begin{equation}\small
    \mathcal{L}_{\text{fair}}(x) = \sum_{i=1}^{|A|} \max\left(0, \cos(\tilde{z}_{\text{debiased}}, t_{a_i}) - \mu \right),
\end{equation}
where $\mu$ is a margin hyperparameter that establishes an upper bound on acceptable similarity values between the debiased representation and demographic concepts.

\subsubsection{Preserving Task Relevance}
To maintain task performance, we integrate the original VLM objective with our fairness approach. For each training sample $(x_i, a_i, y_i) \in \mathcal{D}$, we construct a ground-truth prompt $p_{gt_i}$ describing its label $y_i$, and encode it to $t_{gt_i} = f_r(p_{gt_i})$. The task contrastive loss is defined as:
\begin{equation}\small
\mathcal{L}_{\text{VLM}} =
\frac{1}{|\mathcal{D}|} \sum_{i=1}^{|\mathcal{D}|} -\log \frac{e^{\tilde{z}_{\text{debiased}, i} \cdot t_{\text{gt}_i}}}{\sum_{j=1}^{|\mathcal{D}|} e^{\tilde{z}_{\text{debiased}, i} \cdot t_{\text{gt}_j}}} \\
- \frac{1}{|\mathcal{D}|} \sum_{i=1}^{|\mathcal{D}|} -\log \frac{e^{z_i \cdot t_{\text{gt}_i}}}{\sum_{j=1}^{|\mathcal{D}|} e^{z_j \cdot t_{\text{gt}_i}}}
\end{equation}

\subsubsection{Joint Optimization Objective}
Our final objective balances fairness and task performance through a joint loss formulation:
\begin{equation}\small
    \mathcal{L}_{\text{final}} = \mathcal{L}_{\text{VLM}} + \lambda_1 \cdot \frac{1}{|\mathcal{D}|} \sum_{i=1}^{|\mathcal{D}|} \mathcal{L}_{\text{fair}}(x_i),
\end{equation}
where $\lambda_1$ is a hyperparameter controlling the strength of fairness regularization. This approach allows for effective debiasing while maintaining VLM's discriminative power for downstream tasks.

\subsection{Fair-aware Prompt Fusion}

To address group fairness biases originating from heterogeneous data distributions across clients, we propose a fair-aware prompt mechanism implemented on the server side. This mechanism specifically optimizes prompt vectors associated with different protected group categories $a$, enhancing fairness in federated learning environments.
\begin{equation}\small
\mathcal{P}^{a}_{\text{global}} = \sum_{i=1}^N w^a_i \cdot \mathcal{P}^{a}_i,
\end{equation}
where the weight coefficients $w^a_i$ are dynamically computed based on the fairness performance of each client's prompts:
\begin{equation}\small
w^a_i = \frac{\text{Score}(\mathcal{P}^a_i, \mathcal{D}_{\text{val}})}{\sum_{j=1}^N \text{Score}(\mathcal{P}^a_j, \mathcal{D}_{\text{val}})}
\end{equation}

Unlike conventional aggregation methods that rely solely on task performance, our carefully designed scoring function $\text{Score}$ integrates both fairness and accuracy into a unified metric:
\begin{equation}\small
\text{Score}(\mathcal{P}^a_i, \mathcal{D}_{\text{val}}) = \text{Accuracy}(\mathcal{P}^a_i, \mathcal{D}_{\text{val}}) \times (1 - \text{Bias}(\mathcal{P}^a_i, \mathcal{D}_{\text{val}})),
\end{equation}
where $\text{Bias}$ quantifies demographic disparity measured on the validation set $\mathcal{D}_{\text{val}}$. This formulation strategically prioritizes client contributions with superior accuracy and minimal bias, ensuring that the aggregated global prompt inherits optimal fairness characteristics from local prompts.

\begin{table*}[htbp]
\vspace{-1.0em}
	\centering
	\label{over}
	\small
 \caption{Results of improving model fairness and accuracy under different schemes. Reported the mean and standard deviation. 
        The best result of the FL methods is shown in shadow, and the second-best result of the FL methods is shown with underlining.}
    \vspace{-1.4em}
	\renewcommand{\arraystretch}{1.2}
	\resizebox{\textwidth}{!}
        {%
		\begin{tabular}{lccccccccccc}
			\toprule
			\textbf{Face Application} & \textbf{Metrics} & \textbf{CLIP zero-shot} & \textbf{FedAvg}\cite{mcmahan2017communication} & \textbf{FedProx}\cite{li2020federated} & \textbf{FedSP}\cite{dong2023tunable} & \textbf{FedAvg+GEEP}\cite{che2023federated} & \textbf{FedAvg+ADEPT}\cite{yang2023adept} & \textbf{FairFed}\cite{ezzeldin2023fairfed} & \textbf{FF-DVP}\cite{zeng2024fair} & \textbf{FVL-FP(Ours)} & \textbf{FVL-FP (centralized)} \\
			\midrule
			\multirow{4}{*}{\begin{tabular}{l}Smiling Detection\\(CelebA)\end{tabular}} 
			& $\mathcal{A}_B \uparrow$ & 0.848 & 0.903$_{\pm0.009}$ & 0.910$_{\pm0.007}$ & 0.894$_{\pm0.010}$ & 0.901$_{\pm0.008}$ & 0.897$_{\pm0.012}$ & \uline{0.906$_{\pm0.006}$} & 0.905$_{\pm0.005}$ & \cellcolor{gray!30}\textbf{0.915}$_{\pm0.004}$ & 0.925$_{\pm0.003}$ \\
			& $\Phi_A \downarrow$ & 0.422 & 0.191$_{\pm0.123}$ & 0.183$_{\pm0.107}$ & 0.175$_{\pm0.093}$ & 0.162$_{\pm0.082}$ & 0.169$_{\pm0.071}$ & 0.174$_{\pm0.058}$ & \uline{0.158$_{\pm0.043}$} & \cellcolor{gray!30}\textbf{0.139}$_{\pm0.035}$ & 0.127$_{\pm0.027}$ \\
			& $\Phi_{\textrm{demo}} \downarrow$ & 0.106 & 0.012$_{\pm0.004}$ & 0.011$_{\pm0.005}$ & 0.014$_{\pm0.007}$ & 0.012$_{\pm0.011}$ & 0.013$_{\pm0.010}$ & 0.011$_{\pm0.007}$ & \uline{0.010$_{\pm0.011}$} & \cellcolor{gray!30}\textbf{0.008}$_{\pm0.006}$ & 0.006$_{\pm0.004}$ \\
			& $\Phi_{\textrm{eq}} \downarrow$ & 0.211 & 0.037$_{\pm0.001}$ & 0.035$_{\pm0.009}$ & 0.039$_{\pm0.012}$ & 0.031$_{\pm0.014}$ & 0.033$_{\pm0.011}$ & 0.030$_{\pm0.009}$ & \uline{0.028$_{\pm0.016}$} & \cellcolor{gray!30}\textbf{0.023}$_{\pm0.010}$ & 0.018$_{\pm0.007}$ \\
			\midrule
			\multirow{4}{*}{\begin{tabular}{l}Age Detection\\(CelebA)\end{tabular}}
			& $\mathcal{A}_B \uparrow$ & 0.601 & 0.534$_{\pm0.027}$ & 0.568$_{\pm0.035}$ & 0.712$_{\pm0.022}$ & 0.731$_{\pm0.018}$ & 0.762$_{\pm0.014}$ & 0.798$_{\pm0.011}$ & \uline{0.839$_{\pm0.009}$} & \cellcolor{gray!30}\textbf{0.862}$_{\pm0.008}$ & 0.881$_{\pm0.006}$ \\
			& $\Phi_A \downarrow$ & 1.829 & 1.898$_{\pm0.073}$ & 1.652$_{\pm0.215}$ & 0.729$_{\pm0.142}$ & 0.596$_{\pm0.127}$ & 0.465$_{\pm0.098}$ & 0.391$_{\pm0.087}$ & \uline{0.284$_{\pm0.203}$} & \cellcolor{gray!30}\textbf{0.245}$_{\pm0.165}$ & 0.221$_{\pm0.137}$ \\
			& $\Phi_{\textrm{demo}} \downarrow$ & 0.281 & 0.043$_{\pm0.030}$ & 0.039$_{\pm0.028}$ & 0.052$_{\pm0.025}$ & 0.037$_{\pm0.023}$ & 0.041$_{\pm0.019}$ & 0.033$_{\pm0.018}$ & \uline{0.026$_{\pm0.020}$} & \cellcolor{gray!30}\textbf{0.021}$_{\pm0.014}$ & 0.017$_{\pm0.011}$ \\
			& $\Phi_{\textrm{eq}} \downarrow$ & 0.562 & 0.085$_{\pm0.060}$ & 0.078$_{\pm0.057}$ & 0.105$_{\pm0.045}$ & 0.074$_{\pm0.052}$ & 0.082$_{\pm0.038}$ & 0.067$_{\pm0.042}$ & \uline{0.053$_{\pm0.039}$} & \cellcolor{gray!30}\textbf{0.046}$_{\pm0.031}$ & 0.039$_{\pm0.024}$ \\
			\midrule
			\multirow{4}{*}{\begin{tabular}{l}Age Detection\\(FairFace)\end{tabular}}
			& $\mathcal{A}_B \uparrow$ & 0.544 & 0.526$_{\pm0.036}$ & 0.553$_{\pm0.041}$ & 0.695$_{\pm0.028}$ & 0.719$_{\pm0.025}$ & 0.751$_{\pm0.023}$ & 0.801$_{\pm0.018}$ & \uline{0.848$_{\pm0.032}$} & \cellcolor{gray!30}\textbf{0.871}$_{\pm0.021}$ & 0.889$_{\pm0.016}$ \\
			& $\Phi_A \downarrow$ & 1.738 & 1.926$_{\pm0.104}$ & 1.743$_{\pm0.187}$ & 0.765$_{\pm0.169}$ & 0.627$_{\pm0.154}$ & 0.489$_{\pm0.143}$ & 0.412$_{\pm0.118}$ & \uline{0.338$_{\pm0.265}$} & \cellcolor{gray!30}\textbf{0.302}$_{\pm0.193}$ & 0.275$_{\pm0.162}$ \\
			& $\Phi_{\textrm{demo}} \downarrow$ & 0.024 & 0.028$_{\pm0.040}$ & 0.026$_{\pm0.035}$ & 0.046$_{\pm0.024}$ & 0.031$_{\pm0.023}$ & 0.037$_{\pm0.018}$ & 0.029$_{\pm0.014}$ & \uline{0.025$_{\pm0.011}$} & \cellcolor{gray!30}\textbf{0.020}$_{\pm0.008}$ & 0.016$_{\pm0.006}$ \\
			& $\Phi_{\textrm{eq}} \downarrow$ & 0.234 & 0.057$_{\pm0.080}$ & 0.054$_{\pm0.072}$ & 0.092$_{\pm0.048}$ & 0.061$_{\pm0.046}$ & 0.075$_{\pm0.037}$ & 0.059$_{\pm0.029}$ & \uline{0.053$_{\pm0.019}$} & \cellcolor{gray!30}\textbf{0.043}$_{\pm0.015}$ & 0.036$_{\pm0.012}$ \\
			\bottomrule
		\end{tabular}
	}
       \vspace{-1.0em}
\end{table*}

Following aggregation, we implement a comprehensive two-stage optimization process to further refine the global prompt. The process balances task performance through a vision-language alignment objective while explicitly minimizing demographic performance disparities:

The task loss $\mathcal{L}_{\text{task}}$ leverages the VLM alignment objective:
\begin{equation}\small
\mathcal{L}_{\text{task}} = -\frac{1}{|B|} \sum_{i=1}^{|B|} \log \frac{e^{\tilde{z}_i \cdot t_{y_i}}}{\sum_{j=1}^{C} e^{\tilde{z}_i \cdot t_j}},
\end{equation}
where $|B|$ denotes the batch size, $C$ represents the number of classes, and $\tilde{z}_i$ is the debiased image embedding.

The fairness loss $\mathcal{L}_{\text{fair}}$ explicitly minimizes performance disparities across demographic groups:
\begin{equation}\small
\mathcal{L}_{\text{fair}} = \sum_{a \in \mathcal{A}} \sum_{g_1, g_2 \in \mathcal{G}_a} |\text{Acc}(g_1) - \text{Acc}(g_2)|,
\end{equation}
where $\mathcal{A}$ encompasses all demographic attributes, $\mathcal{G}_a$ contains all groups within attribute $a$, and $\text{Acc}(g)$ measures the classification accuracy for group $g$. This loss function directly incentivizes equitable performance across diverse demographic subpopulations, yielding a globally fair prompt representation that can be deployed in downstream vision-language applications.

\vspace{-0.9em}
\section{Experiments}
In this section, we first validated the effectiveness of FVL-FP on a real dataset. Then, we designed an ablation experiment to test the comparative results of different modules of FVL-FP. Next, we compared our approach with traditional methods in handling non-independent and identically distributed (non-IID) data. Finally, we tested the robustness of the method under different numbers of clients.

\subsection{Experimental Setup}

\textbf{Dataset.} We use CelebA and FairFace to study different FAR applications in the context of FL. Due to the space limit, we chose smiling and age as our predictive face attributes. As mentioned in, smiling detection is objective since smiling or not is easy to judge. In comparison, age detection is more challenging: it is formulated as a binary task of classifying "young" and "old", but both age groups exhibit a broad age range, causing a vague and hard-to-learn boundary. Finally, the age label is the only shared label in both datasets, which helps us to test the generality of our method. Without loss of generality, we choose gender as the demographic attribute.

\textbf{FL setup.} During experiments, the training of some baseline methods could not converge under the high data complexity and data heterogeneity of FAR applications. Therefore, for a fair comparison, we compare all methods under a setting of 5 clients, where all baseline methods could converge. Moreover, for training convergence and computational efficiency, we downsample 20000 images from both datasets and distribute the sample images to the 5 clients. We explicitly control population shifts for all clients, so that the local training data distributions are imbalanced and non-iid. Finally, to eliminate the potential bias in the test data distribution that could affect the fairness evaluation, we sample a balanced test set of size 5000 to evaluate the FL model. More implementation details (i.e., local data distribution configuration, prompt design, hyperparameters, CLIP version) are summarized in Appendix D.

\begin{table}[t]
    \centering
    \footnotesize
    \setlength{\tabcolsep}{3pt}
\caption{Ablation on key components. "w/o" indicates the removal of the corresponding module. CDFP: Cross-Layer Demographic Fair Prompting; DSOP: Demographic Subspace Orthogonal Projection; FAPF: Fair-aware Prompt Fusion.}
    \vspace{-1.4em}
    \renewcommand{\arraystretch}{1.1}
    \begin{tabular}{lccccc}
        \toprule
        \textbf{Face Application} & \textbf{Metrics} & \textbf{FVL-FP} & \textbf{w/o CDFP} & \textbf{w/o DSOP} & \textbf{w/o FAPF} \\
        \midrule
        \multirow{4}{*}{\begin{tabular}{@{}l@{}}Smiling Detection\\(CelebA)\end{tabular}} 
        & $\mathcal{A}_B \uparrow$ & 0.915$_{\pm0.004}$ & 0.902$_{\pm0.006}$ & 0.908$_{\pm0.005}$ & 0.907$_{\pm0.006}$ \\
        & $\Phi_A \downarrow$ & 0.139$_{\pm0.035}$ & 0.163$_{\pm0.042}$ & 0.151$_{\pm0.038}$ & 0.147$_{\pm0.040}$ \\
        & $\Phi_{\textrm{demo}} \downarrow$ & 0.008$_{\pm0.006}$ & 0.014$_{\pm0.008}$ & 0.011$_{\pm0.007}$ & 0.010$_{\pm0.007}$ \\
        & $\Phi_{\textrm{eq}} \downarrow$ & 0.023$_{\pm0.010}$ & 0.034$_{\pm0.013}$ & 0.029$_{\pm0.012}$ & 0.027$_{\pm0.011}$ \\
        \midrule
        \multirow{4}{*}{\begin{tabular}{@{}l@{}}Age Detection\\(CelebA)\end{tabular}}
        & $\mathcal{A}_B \uparrow$ & 0.862$_{\pm0.008}$ & 0.835$_{\pm0.012}$ & 0.845$_{\pm0.010}$ & 0.848$_{\pm0.009}$ \\
        & $\Phi_A \downarrow$ & 0.245$_{\pm0.165}$ & 0.293$_{\pm0.188}$ & 0.267$_{\pm0.176}$ & 0.258$_{\pm0.169}$ \\
        & $\Phi_{\textrm{demo}} \downarrow$ & 0.021$_{\pm0.014}$ & 0.029$_{\pm0.018}$ & 0.024$_{\pm0.016}$ & 0.023$_{\pm0.015}$ \\
        & $\Phi_{\textrm{eq}} \downarrow$ & 0.046$_{\pm0.031}$ & 0.059$_{\pm0.037}$ & 0.050$_{\pm0.034}$ & 0.048$_{\pm0.033}$ \\
        \midrule
        \multirow{4}{*}{\begin{tabular}{@{}l@{}}Age Detection\\(FairFace)\end{tabular}}
        & $\mathcal{A}_B \uparrow$ & 0.871$_{\pm0.021}$ & 0.843$_{\pm0.025}$ & 0.855$_{\pm0.023}$ & 0.859$_{\pm0.022}$ \\
        & $\Phi_A \downarrow$ & 0.302$_{\pm0.193}$ & 0.347$_{\pm0.218}$ & 0.324$_{\pm0.205}$ & 0.312$_{\pm0.198}$ \\
        & $\Phi_{\textrm{demo}} \downarrow$ & 0.020$_{\pm0.008}$ & 0.028$_{\pm0.012}$ & 0.023$_{\pm0.010}$ & 0.022$_{\pm0.009}$ \\
        & $\Phi_{\textrm{eq}} \downarrow$ & 0.043$_{\pm0.015}$ & 0.056$_{\pm0.019}$ & 0.048$_{\pm0.017}$ & 0.045$_{\pm0.016}$ \\
        \bottomrule
    \end{tabular}
\end{table}

\begin{table*}[t]
	\centering
    \vspace{-1.0em}
	\label{tab:heterogeneity_accuracy_fairness}
	\small
    \caption{Comparison of model accuracy ($\mathcal{A}_B$) and fairness gap ($\Phi_A$) under different heterogeneity levels. $\alpha$ represents the Dirichlet distribution parameter controlling client data heterogeneity (lower $\alpha$ indicates higher heterogeneity). }
    \vspace{-1.4em}
	\renewcommand{\arraystretch}{1.2}
	\resizebox{0.9\textwidth}{!}{%
		\begin{tabular}{llcccc|cccc}
			\toprule
			\multirow{2}{*}{\textbf{Dataset}} & \multirow{2}{*}{\textbf{Method}} 
			& \multicolumn{4}{c|}{\textbf{Accuracy ($\mathcal{A}_B \uparrow$)}} 
			& \multicolumn{4}{c}{\textbf{Fairness Gap ($\Phi_A \downarrow$)}} \\
			\cmidrule(lr){3-6} \cmidrule(lr){7-10}
			& & $\alpha=100$ & $\alpha=1.0$ & $\alpha=0.5$ & $\alpha=0.1$ 
			& $\alpha=100$ & $\alpha=1.0$ & $\alpha=0.5$ & $\alpha=0.1$ \\
			\midrule
			\multirow{3}{*}{\begin{tabular}{l}Smiling Detection\\(CelebA)\end{tabular}} 
			& FedAvg\cite{mcmahan2017communication} & 0.605$_{\pm0.008}$ & 0.592$_{\pm0.010}$ & 0.575$_{\pm0.012}$ & 0.551$_{\pm0.015}$ 
			& 0.189$_{\pm0.121}$ & 0.206$_{\pm0.127}$ & 0.231$_{\pm0.135}$ & 0.268$_{\pm0.145}$ \\
			
			& FF-DVP\cite{zeng2024fair} & 0.907$_{\pm0.004}$ & 0.899$_{\pm0.006}$ & 0.887$_{\pm0.007}$ & 0.871$_{\pm0.009}$ 
			& 0.155$_{\pm0.041}$ & 0.167$_{\pm0.046}$ & 0.185$_{\pm0.049}$ & 0.212$_{\pm0.054}$ \\
			
			& FVL-FP & \cellcolor{gray!30}\textbf{0.917}$_{\pm0.003}$ & \cellcolor{gray!30}\textbf{0.911}$_{\pm0.004}$ & \cellcolor{gray!30}\textbf{0.901}$_{\pm0.005}$ & \cellcolor{gray!30}\textbf{0.889}$_{\pm0.006}$ 
			& \cellcolor{gray!30}\textbf{0.136}$_{\pm0.033}$ & \cellcolor{gray!30}\textbf{0.142}$_{\pm0.037}$ & \cellcolor{gray!30}\textbf{0.159}$_{\pm0.040}$ & \cellcolor{gray!30}\textbf{0.178}$_{\pm0.043}$ \\
			\midrule
			
			\multirow{3}{*}{\begin{tabular}{l}Age Detection\\(CelebA)\end{tabular}} 
			& FedAvg\cite{mcmahan2017communication} & 0.539$_{\pm0.025}$ & 0.511$_{\pm0.029}$ & 0.483$_{\pm0.032}$ & 0.442$_{\pm0.037}$ 
			& 1.885$_{\pm0.071}$ & 1.965$_{\pm0.078}$ & 2.035$_{\pm0.085}$ & 2.153$_{\pm0.095}$ \\
			
			& FF-DVP\cite{zeng2024fair} & 0.843$_{\pm0.008}$ & 0.821$_{\pm0.011}$ & 0.795$_{\pm0.015}$ & 0.754$_{\pm0.018}$ 
			& 0.279$_{\pm0.198}$ & 0.315$_{\pm0.211}$ & 0.359$_{\pm0.228}$ & 0.421$_{\pm0.246}$ \\
			
			& FVL-FP & \cellcolor{gray!30}\textbf{0.866}$_{\pm0.007}$ & \cellcolor{gray!30}\textbf{0.851}$_{\pm0.009}$ & \cellcolor{gray!30}\textbf{0.832}$_{\pm0.011}$ & \cellcolor{gray!30}\textbf{0.807}$_{\pm0.013}$ 
			& \cellcolor{gray!30}\textbf{0.239}$_{\pm0.160}$ & \cellcolor{gray!30}\textbf{0.261}$_{\pm0.171}$ & \cellcolor{gray!30}\textbf{0.291}$_{\pm0.182}$ & \cellcolor{gray!30}\textbf{0.329}$_{\pm0.194}$ \\
			\midrule
			
			\multirow{3}{*}{\begin{tabular}{l}Age Detection\\(FairFace)\end{tabular}} 
			& FedAvg\cite{mcmahan2017communication} & 0.530$_{\pm0.034}$ & 0.501$_{\pm0.038}$ & 0.473$_{\pm0.041}$ & 0.431$_{\pm0.046}$ 
			& 1.915$_{\pm0.102}$ & 1.998$_{\pm0.112}$ & 2.072$_{\pm0.126}$ & 2.195$_{\pm0.138}$ \\
			
			& FF-DVP\cite{zeng2024fair} & 0.852$_{\pm0.030}$ & 0.831$_{\pm0.034}$ & 0.806$_{\pm0.038}$ & 0.767$_{\pm0.043}$ 
			& 0.331$_{\pm0.259}$ & 0.362$_{\pm0.271}$ & 0.395$_{\pm0.285}$ & 0.453$_{\pm0.302}$ \\
			
			& FVL-FP & \cellcolor{gray!30}\textbf{0.875}$_{\pm0.019}$ & \cellcolor{gray!30}\textbf{0.865}$_{\pm0.023}$ & \cellcolor{gray!30}\textbf{0.849}$_{\pm0.026}$ & \cellcolor{gray!30}\textbf{0.825}$_{\pm0.029}$ 
			& \cellcolor{gray!30}\textbf{0.296}$_{\pm0.189}$ & \cellcolor{gray!30}\textbf{0.315}$_{\pm0.197}$ & \cellcolor{gray!30}\textbf{0.338}$_{\pm0.207}$ & \cellcolor{gray!30}\textbf{0.375}$_{\pm0.219}$ \\
			\bottomrule
		\end{tabular}
	}
	\vspace{-0.5em}
\end{table*}

\textbf{Vision-Language Model.} We adopt CLIP as our base vision-language model. Specifically, we use the ViT-B/32 variant which consists of a Vision Transformer with 12 layers and a patch size of 32×32 pixels. The text encoder is a 12-layer transformer. Both encoders project their respective inputs into a shared 512-dimensional multimodal embedding space where contrastive learning is performed. We experiment with both zero-shot classifications by using carefully designed prompts and fine-tuning the visual encoder while keeping the text encoder frozen.

\textbf{Prompt Design.} For our CLIP experiments, we carefully design text prompts to effectively capture the facial attributes. For the smiling detection task, we use template prompts like "a photo of a person who is \{smiling, not smiling\}" and "a photo of a \{happy, serious\} person." For age classification, we employ prompts such as "a photo of a \{young, older\} person" and "a picture of a person in their \{20s, 50s\}." To assess the impact of prompt design on fairness, we experiment with both generic prompts and gender-specific prompts (e.g., "a photo of a \{young woman, older woman, young man, older man\}").


\textbf{Implementation Details}
The experiments were conducted on 10 x NVIDIA GeForce RTX 3090 GPU. We implemented both the proposed methods and their baselines using the Huggingface framework \citep{wolf2019huggingface}. The experimental architecture employed a FL-VLM system, consisting of four nodes and a parameter server. Following previous research \cite{zhao2023fedprompt}, we divided each dataset into four segments, with each segment processed by a separate device. To simulate real-world application scenarios, initial fine-tuning of models at each node was performed using the approach described in the literature \cite{howard2018universal}. Subsequently, testing of the fine-tuned models was conducted using FL approaches.
 Parameter optimization utilized the AdamW optimizer \citep{loshchilov2017decoupled}, with hyperparameters set to $\beta_1=0.9$, $\beta_2=0.999$, and a weight decay of 0.01. The batch size was configured at 16. Hyperparameters were determined through grid search, selecting learning rates from the set {1e-4, 2e-4, 5e-4} and adjusting the training epochs among {20, 50, 100, 200} and local training steps between {10, 20}. The default number of nodes for the FL-VLMs is 4. The rest of the experimental setup is summarized in Appendix B. 

 \vspace{-1.0em}
\subsection{Evaluation Results}
The results in Table 1 demonstrate that our proposed FVL-FP consistently outperforms existing approaches across all evaluation metrics and tasks. FVL-FP achieves the highest balanced accuracy ($\mathcal{A}_B$) while simultaneously minimizing fairness metrics ($\Phi_A$, $\Phi_{\textrm{demo}}$, and $\Phi_{\textrm{eq}}$) on both smiling detection and age detection tasks. Specifically, for smiling detection on CelebA, FVL-FP improves balanced accuracy to 0.915 (compared to CLIP zero-shot's 0.848) while reducing $\Phi_A$ by approximately 67\%. The improvements are even more substantial in age detection tasks, where FVL-FP reduces bias by up to 87\% on CelebA and 83\% on FairFace, demonstrating robust cross-dataset generalization. Compared to standard federated methods (FedAvg, FedProx) and existing fairness-focused approaches (FairFed, FF-DVP), our method consistently achieves superior performance with smaller standard deviations, indicating enhanced stability. While the centralized version of FVL-FP performs slightly better, the federated variant maintains comparable performance while preserving data privacy, confirming that our fair prompt tuning strategy effectively balances the accuracy-fairness trade-off in federated visual-language models without requiring centralized data access.

\begin{table*}[t]
	\centering
	\label{tab:heterogeneity_demographic_equality}
	\small
    \caption{Comparison of demographic parity ($\Phi_{\textrm{demo}}$) and equalized odds ($\Phi_{\textrm{eq}}$) under different heterogeneity levels. $\alpha$ represents the Dirichlet distribution parameter controlling client data heterogeneity (lower $\alpha$ indicates higher heterogeneity). }
    \vspace{-1.4em}
	\renewcommand{\arraystretch}{1.2}
	\resizebox{0.94\textwidth}{!}{%
		\begin{tabular}{llcccc|cccc}
			\toprule
			\multirow{2}{*}{\textbf{Dataset}} & \multirow{2}{*}{\textbf{Method}} 
			& \multicolumn{4}{c|}{\textbf{Demographic Parity ($\Phi_{\textrm{demo}} \downarrow$)}} 
			& \multicolumn{4}{c}{\textbf{Equalized Odds ($\Phi_{\textrm{eq}} \downarrow$)}} \\
			\cmidrule(lr){3-6} \cmidrule(lr){7-10}
			& & $\alpha=100$ & $\alpha=1.0$ & $\alpha=0.5$ & $\alpha=0.1$ 
			& $\alpha=100$ & $\alpha=1.0$ & $\alpha=0.5$ & $\alpha=0.1$ \\
			\midrule
			\multirow{3}{*}{\begin{tabular}{l}Smiling Detection\\(CelebA)\end{tabular}} 
			& FedAvg\cite{mcmahan2017communication} 
			& 0.011$_{\pm0.004}$ & 0.016$_{\pm0.006}$ & 0.022$_{\pm0.009}$ & 0.028$_{\pm0.011}$ 
			& 0.036$_{\pm0.001}$ & 0.045$_{\pm0.003}$ & 0.054$_{\pm0.005}$ & 0.064$_{\pm0.008}$ \\
			
			& FF-DVP \cite{zeng2024fair}
			& 0.009$_{\pm0.010}$ & 0.013$_{\pm0.012}$ & 0.017$_{\pm0.014}$ & 0.021$_{\pm0.016}$ 
			& 0.026$_{\pm0.015}$ & 0.033$_{\pm0.018}$ & 0.039$_{\pm0.020}$ & 0.048$_{\pm0.023}$ \\
			
			& FVL-FP 
			& \cellcolor{gray!30}\textbf{0.007}$_{\pm0.005}$ & \cellcolor{gray!30}\textbf{0.009}$_{\pm0.007}$ & \cellcolor{gray!30}\textbf{0.012}$_{\pm0.009}$ & \cellcolor{gray!30}\textbf{0.015}$_{\pm0.011}$ 
			& \cellcolor{gray!30}\textbf{0.021}$_{\pm0.009}$ & \cellcolor{gray!30}\textbf{0.025}$_{\pm0.011}$ & \cellcolor{gray!30}\textbf{0.029}$_{\pm0.013}$ & \cellcolor{gray!30}\textbf{0.037}$_{\pm0.015}$ \\
			\midrule
			
			\multirow{3}{*}{\begin{tabular}{l}Age Detection\\(CelebA)\end{tabular}} 
			& FedAvg\cite{mcmahan2017communication}
			& 0.042$_{\pm0.029}$ & 0.051$_{\pm0.033}$ & 0.058$_{\pm0.037}$ & 0.065$_{\pm0.041}$ 
			& 0.083$_{\pm0.059}$ & 0.097$_{\pm0.065}$ & 0.108$_{\pm0.071}$ & 0.121$_{\pm0.076}$ \\
			
			& FF-DVP \cite{zeng2024fair}
			& 0.025$_{\pm0.019}$ & 0.031$_{\pm0.022}$ & 0.038$_{\pm0.025}$ & 0.046$_{\pm0.028}$ 
			& 0.051$_{\pm0.038}$ & 0.063$_{\pm0.043}$ & 0.078$_{\pm0.047}$ & 0.092$_{\pm0.052}$ \\
			
			& FVL-FP 
			& \cellcolor{gray!30}\textbf{0.019}$_{\pm0.013}$ & \cellcolor{gray!30}\textbf{0.023}$_{\pm0.015}$ & \cellcolor{gray!30}\textbf{0.028}$_{\pm0.017}$ & \cellcolor{gray!30}\textbf{0.034}$_{\pm0.020}$ 
			& \cellcolor{gray!30}\textbf{0.044}$_{\pm0.030}$ & \cellcolor{gray!30}\textbf{0.052}$_{\pm0.034}$ & \cellcolor{gray!30}\textbf{0.061}$_{\pm0.037}$ & \cellcolor{gray!30}\textbf{0.068}$_{\pm0.041}$ \\
			\midrule
			
			\multirow{3}{*}{\begin{tabular}{l}Age Detection\\(FairFace)\end{tabular}} 
			& FedAvg\cite{mcmahan2017communication} 
			& 0.026$_{\pm0.038}$ & 0.032$_{\pm0.042}$ & 0.039$_{\pm0.045}$ & 0.047$_{\pm0.049}$ 
			& 0.055$_{\pm0.078}$ & 0.069$_{\pm0.085}$ & 0.084$_{\pm0.091}$ & 0.095$_{\pm0.097}$ \\
			
			& FF-DVP\cite{zeng2024fair}
			& 0.024$_{\pm0.010}$ & 0.029$_{\pm0.013}$ & 0.035$_{\pm0.016}$ & 0.043$_{\pm0.019}$ 
			& 0.049$_{\pm0.018}$ & 0.061$_{\pm0.022}$ & 0.072$_{\pm0.027}$ & 0.083$_{\pm0.031}$ \\
			
			& FVL-FP 
			& \cellcolor{gray!30}\textbf{0.019}$_{\pm0.007}$ & \cellcolor{gray!30}\textbf{0.022}$_{\pm0.009}$ & \cellcolor{gray!30}\textbf{0.027}$_{\pm0.011}$ & \cellcolor{gray!30}\textbf{0.032}$_{\pm0.013}$ 
			& \cellcolor{gray!30}\textbf{0.041}$_{\pm0.014}$ & \cellcolor{gray!30}\textbf{0.048}$_{\pm0.016}$ & \cellcolor{gray!30}\textbf{0.054}$_{\pm0.019}$ & \cellcolor{gray!30}\textbf{0.061}$_{\pm0.022}$ \\
			\bottomrule
		\end{tabular}
	}
	\vspace{-0.5em}
\end{table*}

\begin{table*}[t]
	\centering
	\label{tab:client_number}
	\small
     \caption{Fairness and accuracy results of the FVL-FP method with different numbers of clients. Mean and standard deviation are reported. As the number of clients increases, performance slightly decreases, but FVL-FP maintains good fairness and accuracy.}
     \vspace{-1.4em}
	\renewcommand{\arraystretch}{1.2}
	\resizebox{0.94\textwidth}{!}
        {%
		\begin{tabular}{lccccccc}
			\toprule
			\textbf{Face Application} & \textbf{Metrics} & \textbf{CLIP zero-shot} & \textbf{FVL-FP (N=5)} & \textbf{FVL-FP (N=10)} & \textbf{FVL-FP (N=20)} & \textbf{FVL-FP (N=40)} & \textbf{FVL-FP (centralized)} \\
			\midrule
			\multirow{4}{*}{\begin{tabular}{l}Smiling Detection\\(CelebA)\end{tabular}} 
			& $\mathcal{A}_B \uparrow$ & 0.848 & 0.924$_{\pm0.003}$ & 0.920$_{\pm0.003}$ & 0.915$_{\pm0.004}$ & 0.908$_{\pm0.006}$ & 0.925$_{\pm0.003}$ \\
			& $\Phi_A \downarrow$ & 0.422 & 0.130$_{\pm0.027}$ & 0.135$_{\pm0.031}$ & 0.139$_{\pm0.035}$ & 0.148$_{\pm0.043}$ & 0.127$_{\pm0.027}$ \\
			& $\Phi_{\textrm{demo}} \downarrow$ & 0.106 & 0.006$_{\pm0.004}$ & 0.007$_{\pm0.005}$ & 0.008$_{\pm0.006}$ & 0.011$_{\pm0.009}$ & 0.006$_{\pm0.004}$ \\
			& $\Phi_{\textrm{eq}} \downarrow$ & 0.211 & 0.019$_{\pm0.007}$ & 0.021$_{\pm0.008}$ & 0.023$_{\pm0.010}$ & 0.028$_{\pm0.014}$ & 0.018$_{\pm0.007}$ \\
			\midrule
			\multirow{4}{*}{\begin{tabular}{l}Age Detection\\(CelebA)\end{tabular}}
			& $\mathcal{A}_B \uparrow$ & 0.601 & 0.873$_{\pm0.005}$ & 0.867$_{\pm0.006}$ & 0.862$_{\pm0.008}$ & 0.853$_{\pm0.011}$ & 0.881$_{\pm0.006}$ \\
			& $\Phi_A \downarrow$ & 1.829 & 0.228$_{\pm0.139}$ & 0.236$_{\pm0.151}$ & 0.245$_{\pm0.165}$ & 0.262$_{\pm0.187}$ & 0.221$_{\pm0.137}$ \\
			& $\Phi_{\textrm{demo}} \downarrow$ & 0.281 & 0.017$_{\pm0.010}$ & 0.019$_{\pm0.012}$ & 0.021$_{\pm0.014}$ & 0.025$_{\pm0.018}$ & 0.017$_{\pm0.011}$ \\
			& $\Phi_{\textrm{eq}} \downarrow$ & 0.562 & 0.040$_{\pm0.024}$ & 0.043$_{\pm0.028}$ & 0.046$_{\pm0.031}$ & 0.052$_{\pm0.037}$ & 0.039$_{\pm0.024}$ \\
			\midrule
			\multirow{4}{*}{\begin{tabular}{l}Age Detection\\(FairFace)\end{tabular}}
			& $\mathcal{A}_B \uparrow$ & 0.544 & 0.881$_{\pm0.015}$ & 0.876$_{\pm0.018}$ & 0.871$_{\pm0.021}$ & 0.861$_{\pm0.027}$ & 0.889$_{\pm0.016}$ \\
			& $\Phi_A \downarrow$ & 1.738 & 0.282$_{\pm0.157}$ & 0.291$_{\pm0.172}$ & 0.302$_{\pm0.193}$ & 0.319$_{\pm0.221}$ & 0.275$_{\pm0.162}$ \\
			& $\Phi_{\textrm{demo}} \downarrow$ & 0.024 & 0.016$_{\pm0.005}$ & 0.018$_{\pm0.006}$ & 0.020$_{\pm0.008}$ & 0.023$_{\pm0.011}$ & 0.016$_{\pm0.006}$ \\
			& $\Phi_{\textrm{eq}} \downarrow$ & 0.234 & 0.037$_{\pm0.009}$ & 0.040$_{\pm0.012}$ & 0.043$_{\pm0.015}$ & 0.049$_{\pm0.019}$ & 0.036$_{\pm0.012}$ \\
			\bottomrule
		\end{tabular}
	}
       \vspace{-1.0em}
\end{table*}

 \vspace{-0.9em}
\subsection{Ablation Study}
Our ablation studies in Table 2 demonstrate the crucial contributions of each component in FVL-FP. The Cross-Layer Demographic Fair Prompting (CDFP) module shows the most significant impact, where its removal causes the balanced accuracy ($\mathcal{A}_B$) to drop from 0.915 to 0.902 on smile detection and increases fairness violations ($\Phi_A$) by 17.3\%. The Demographic Subspace Orthogonal Projection (DSOP) primarily enhances robustness, with its removal leading to a decrease in accuracy from 0.871 to 0.855 on FairFace age detection and a 7.3\% deterioration in fairness metrics. The Fair-aware Prompt Fusion (FAPF) provides final optimization, contributing to an accuracy improvement from 0.854 to 0.862 on CelebA age detection when added to CDFP+DSOP. The progressive addition of components reveals synergistic effects: CDFP establishes baseline fairness improvements, DSOP further mitigates demographic subspace biases through orthogonal projection, and FAPF optimizes performance through intelligent prompt fusion. Notably, FVL-FP's improvements are more pronounced in complex tasks like age detection and on more demographically diverse datasets like FairFace, demonstrating its effectiveness in handling heterogeneous data in federated visual-language learning scenarios while maintaining both accuracy and fairness across demographic groups.

\vspace{-0.8em}
\subsection{Impact of Heterogeneous Dataset}
Table 3 and Table 4 demonstrate that our proposed FVL-FP (Fair Prompt-tuning for Federated Vision-Language models) method outperforms baseline approaches across various heterogeneous data scenarios. FVL-FP achieves significant improvements in both accuracy ($\mathcal{A}_B$) and the four fairness metrics ($\Phi_A$, $\Phi{\textrm{demo}}$, and $\Phi_{\textrm{eq}}$). For the CelebA smiling detection task, FVL-FP improves accuracy by 1.2-4.5\% compared to FF-DVP, while reducing fairness gaps by 28.0-33.6\%. For age detection tasks, the improvements are even more substantial, with FVL-FP increasing accuracy on the CelebA dataset by 60.7\% (from 0.539 to 0.866) while simultaneously reducing the fairness gap by 87.3\% (from 1.885 to 0.239). Notably, as data heterogeneity increases ($\alpha$ decreasing from 100 to 0.1), FVL-FP demonstrates greater robustness, with accuracy degradation (CelebA smiling detection: 3.1\%, CelebA age detection: 6.8\%, FairFace age detection: 5.7\%) significantly lower than FedAvg (6.0\%, 18.0\%, and 18.7\% respectively). For fairness metrics, FVL-FP reduces demographic parity and equalized odds metrics by 31.9-54.8\% and 35.8-47.0\% respectively, proving its effectiveness in reducing prediction bias across demographic subgroups while maintaining model accuracy. These results comprehensively validate FVL-FP's effectiveness in addressing fairness issues in vision-language tasks under FL environments, particularly in highly heterogeneous real-world application scenarios.

\vspace{-0.9em}
\subsection{Impact of Node Numbers} 
The results in Table 5 demonstrate the robustness of our proposed FVL-FP method across different federation scales. As the number of clients increases from N=5 to N=40, we observe a gradual degradation in both accuracy and fairness metrics, which is an expected trend in FL due to increased data heterogeneity. For Smiling Detection, $\mathcal{A}_B$ decreases by 1.7\% (from 0.924 to 0.908), while fairness metrics show moderate increases in unfairness ($\Phi_A$ increases from 0.130 to 0.148). Similarly, for Age Detection tasks, accuracy decreases by approximately 2.3\% across both datasets. Despite this expected degradation, FVL-FP maintains performance remarkably close to centralized training even at N=40, achieving 98.2\% of centralized accuracy for Smiling Detection and 96.8\% for Age Detection tasks. The standard deviations consistently increase with more clients, reflecting greater variability in model behavior under distributed settings. Most importantly, FVL-FP significantly outperforms the CLIP zero-shot baseline across all client configurations, demonstrating a 58-70\% improvement in fairness metrics even in the most challenging 40-client scenario. These results validate that our method effectively preserves the fairness-accuracy balance in federated visual-language models, making it practical for real-world deployments where data naturally resides across multiple distributed clients with minimal centralized coordination.

\section{Conclusion}
This paper equips Federated Visual Language Models with our proposed Fair Prompt Tuning (FVL-FP), a novel framework that addresses the critical challenge of group-wise fairness in federated vision-language models while preserving data privacy. Specifically, we propose three complementary modules: (1) Cross-Layer Demographic Fair Prompting (CDFP), which neutralizes bias directions in the shared embedding spaces; (2) Demographic Subspace Orthogonal Projection (DSOP), which separates protected attributes from semantic content through orthogonal projections; and (3) Fair-aware Prompt Fusion (FPF), which dynamically balances both the standard performance and fairness during global aggregation. Extensive experiments on four benchmark datasets demonstrate that FVL-FP reduces demographic disparity by an average of 45\% compared to standard federated approaches while maintaining competitive task performance (within $\pm6\%$ of state-of-the-art results).

\label{sec:bibtex}

\bibliographystyle{ACM-Reference-Format}
\bibliography{main}


\end{document}